\documentclass[letterpaper, 10 pt, conference]{ieeeconf}  %

\IEEEoverridecommandlockouts    %
\usepackage{graphics} %
\usepackage{graphicx}
\usepackage{epsfig} %
\usepackage{times} %
\usepackage{amsmath} %
\usepackage{amssymb}  %
\usepackage{float}

\usepackage{booktabs}

\usepackage{xcolor}
\usepackage{colortbl}

\usepackage{pifont}%

\newcommand{\cmark}{\ding{51}}%
\newcommand{\xmark}{\ding{55}}%

\usepackage{siunitx}  
\usepackage{subcaption}
\usepackage{lipsum} %

\usepackage[colorlinks = true,
            linkcolor = blue,
            urlcolor  = blue,
            citecolor = blue,
            anchorcolor = blue]{hyperref}

\usepackage{tikz}
\usepackage{leftidx}

\usepackage{bm}

\usepackage[normalem]{ulem} %
\usetikzlibrary{shapes.geometric, arrows}

\tikzstyle{startstop} = [rectangle, rounded corners, minimum width=3cm, minimum height=0.1cm,text centered, draw=black]
\tikzstyle{arrow} = [thick,->,>=stealth]

\usepackage{bm}
\usepackage{siunitx} 

\usepackage{mathtools}

\usepackage{mathtools}

\usepackage[utf8]{inputenc}

\usepackage{tikz}
\usepackage{tikz-cd}
\usepackage{pgfplots}
\pgfplotsset{compat=1.14}
\usepackage{ulem}
\usepackage{acro}
\title{\LARGE \bf FOCI: Trajectory Optimization on
Gaussian Splats}

\author{Mario Gomez Andreu*$^{1}$, Maximum Wilder-Smith*$^{1}$, Victor Klemm$^{1}$, \\ Vaishakh Patil$^{1}$, Jesus Tordesillas$^{2}$, Marco Hutter$^{1}$
\thanks{*These authors contributed equally}
\thanks{$^{1}$The authors are with the Robotic Systems Lab, ETH Z\"urich, Switzerland.\tt{ \{margomez, mwilder, vklemm, patilv, mahutter \}@ethz.ch}}
\thanks{$^{2}$The author is with the IIT and ICAI, Comillas Pontifical University (Spain). \tt{jtordesillas@comillas.edu}}
\thanks{This work was funded by NCCR Automation, and Swiss Federal Railways (SBB) via ETH Mobility Initiative.}}

\DeclareAcronym{3dgs}{
short = 3DGS ,
long = 3D Gaussian Splatting,
sort = 3DGaussianSplat
}

\DeclareAcronym{nerf}{
short = NeRF ,
long = Neural Radiance Field,
short-plural = s ,
long-plural = s ,
sort = NeuralRadianceField
}

\DeclareAcronym{esdf}{
short = ESDF ,
long = Euclidean Signed Distance Field,
short-plural = s ,
long-plural = s ,
sort = EuclideanSignedDistanceField
}

\begin{document}

\maketitle
\thispagestyle{empty}
\pagestyle{empty}

\begin{abstract}
\ac{3dgs} has recently gained popularity as a faster alternative to \acp{nerf} in 3D reconstruction and view synthesis methods. 
Leveraging the spatial information encoded in \ac{3dgs}, 
this work proposes FOCI (Field Overlap Collision Integral), an algorithm that is able to optimize trajectories directly on the Gaussians themselves. FOCI leverages a novel and interpretable collision formulation for \ac{3dgs} using the notion of the overlap integral between Gaussians. 
Contrary to other approaches, which represent the robot with conservative bounding boxes that underestimate the traversability of the environment, we propose to represent the environment \textit{and} the robot as Gaussian Splats. This not only has desirable computational properties, but also allows for orientation-aware planning, allowing the robot to pass through very tight and narrow spaces.    
We extensively test our algorithm in both synthetic and real Gaussian Splats, showcasing that collision-free trajectories for the ANYmal legged robot that can be computed in a few seconds, even with hundreds of thousands of Gaussians making up the environment. 
The project page and code are available at \href{https://rffr.leggedrobotics.com/works/foci}
{https://rffr.leggedrobotics.com/works/foci/}

\end{abstract}

\section{INTRODUCTION}
Trajectory planning is integral to autonomous mobile robotics to ensure guided and safe operation.
However, in order to make an informed decision, these planning algorithms heavily depend on the underlying environment representations.
Popular representations include occupancy grids, signed distance fields, 3D Meshes, and point clouds. 

Recently, \acp{nerf}~\cite{Mildenhall2020}  have been proposed as a novel neural representation of the environment. They can be created from simple monocular images and they encode the environment as a fully connected neural network, mapping position in space and viewing direction to occupancy and color. However, they suffer from having slow inference speeds because new views have to be created using a computationally expensive ray-casting procedure. More recently, \ac{3dgs}~\cite{Kerbl2023} has been proposed as a promising alternative to \acp{nerf}. Instead of using a neural network to represent the radiance field, it is defined explicitly by a set of 3D Gaussian ellipsoids. Because of this, new views can be efficiently generated by projecting these ellipsoids into the view plane instead of sampling the implicit density of a \ac{nerf} along a ray. Not only does this accelerate novel view synthesis on learned scenes, but the rendering speeds allow for rapid training of environments in a matter of minutes.  

Given the advantages that \ac{3dgs} offers when compared with \acp{nerf}, the natural question is how a robot can leverage this Gaussian representation for navigation. \textbf{In this paper, we propose an algorithm that enables a robot to perform trajectory optimization directly on the 3D Gaussians.} Although some steps have been taken in this direction~\cite{Chen2024,lei2024gaussnav,goel2024distance}, the huge number of Gaussians a scene can have, together with the specific formulation of an explicit collision measure, makes this problem especially hard.  

\begin{figure}[t]
    \centering
    \includegraphics[width=\textwidth/2]{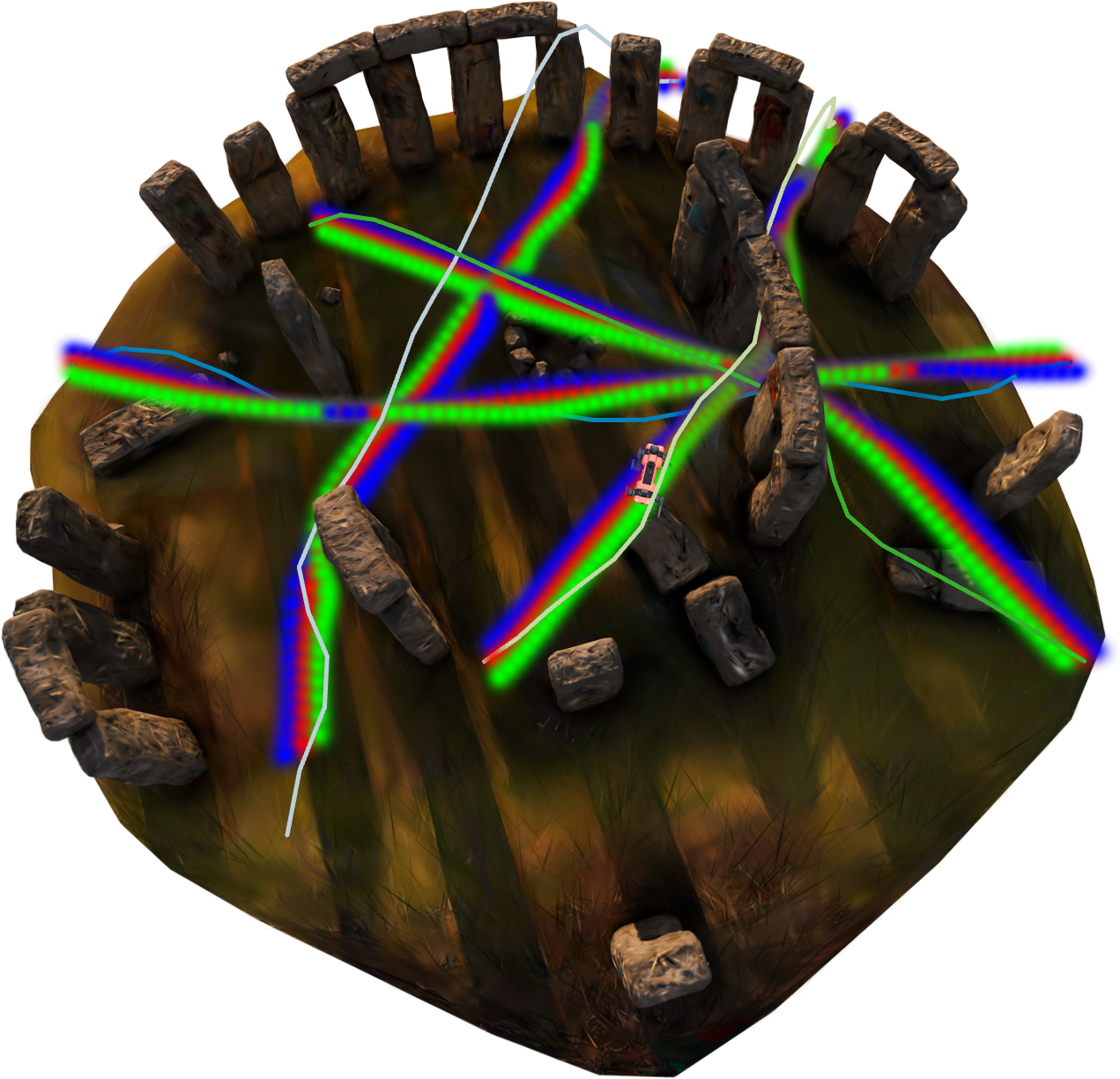}
    \caption{Four different orientation-aware trajectories planned through a Stonehenge environment for an ANYmal quadrupted robot~\cite{hutter2016anymal} . Splines show A* initial trajectories while green, red and blue Gaussians show an orientation-aware plan with \textcolor{green}{green} representing the front, \textcolor{red}{red} the center, and \textcolor{blue}{blue} the rear.}
    \label{fig:ANYmal_title}
\end{figure}

\begin{table*}[!htbp]
\centering
\begin{tabular}{|l|c|c|c|c|c|}
\hline
& \textbf{Splat-Nav \cite{Chen2024}} & \textbf{GS-Planner \cite{jin2024gs}} & \textbf{GSM \cite{goel2024distance}} & \textbf{GaussNav \cite{lei2024gaussnav}} & \textbf{FOCI (ours)} \\
\hline
Planning pipeline included & \textcolor{green}{\cmark} & \textcolor{green}{\cmark} & \textcolor{red}{\xmark} & \textcolor{green}{\cmark} & \textcolor{green}{\cmark} \\
\hline
Orientation-aware collision & \textcolor{red}{\xmark} & \textcolor{red}{\xmark} & \textcolor{red}{\xmark} & \textcolor{red}{\xmark} & \textcolor{green}{\cmark} \\
\hline
Covariance information leveraged & \textcolor{green}{\cmark} & \textcolor{green}{\cmark} & \textcolor{green}{\cmark} & \textcolor{red}{\xmark} & \textcolor{green}{\cmark} \\
\hline
\end{tabular}
\caption{Comparison of our work with related works on Gaussian Splats}
\label{tab:comparison}
\end{table*}

To overcome these challenges, we propose \textbf{FOCI}, a trajectory optimization algorithm that leverages the overlap integral - the spatial integral over the multiplication of two functions - as a proxy measure for the collision between two Gaussians. By representing both the robot \textit{and} the environment with 3D Gaussians, evaluating the full-body collision between them reduces to a sum of normal distribution evaluations.
Furthermore, the resulting expression is fully differentiable, yielding expressive gradients in the optimization step.

The contributions of this work are therefore summarized as follows:
\begin{itemize}
    \item A novel collision measure between Gaussian Splats based on the overlap integral between Gaussians.
    \item A trajectory planning algorithm that uses this formulation to optimize fully differentiable trajectories.
    \item A GPU implementation of the trajectory optimization. 
    \item An evaluation of the proposed algorithm on the ANYmal legged robot in simulation and the real world.
\end{itemize}

\section{Background and Related Work}

Environmental representations are essential parts of the planning pipeline. They determine what methods can be used for trajectory generation, as well as what sensors and information are needed for real-world deployments.

\subsection{Radiance Fields}
Radiance Fields are widely used in visual computing to represent 3D scenes and perform novel view synthesis, creating photorealistic renders from out of domain poses.
The preliminary form is a \acl{nerf}, initially proposed by Mildenhall et al.~\cite{mildenhall2020NeRF}, which are a type of environment representation that provide a neural mapping from position in space and a viewing angle to visual density and color along a ray. To render an image this neural radiance is integrated along a series of rays capturing view-dependent effects like specularity and volumetrics. \\

However, this ray casting operation can be costly at scale, and the implicit scene representation can make direct sampling difficult. As an alternative, \ac{3dgs}~\cite{Kerbl2023} produces similar quality reconstruction with an explicit basis at far faster speeds. Instead of encoding the scene in a neural network, \ac{3dgs} creates a series of 3D Gaussian ellipsoids with colors and Spherical Harmonics~\cite{plenoxels} to encode view-dependent effects. Instead of ray casting each Gaussian, the ellipsoids are splatted~\cite{964490} onto the viewing plane for rasterization levels of speed.
Similar to \acp{nerf}, \ac{3dgs} is optimized by minimizing the deviation between an actual image and the corresponding image predicted by the model.
Because of their explicit nature, \ac{3dgs} allows for an informed guess that initializes all Gaussian means to points from Structure from Motion.
The photometric losses then optimize the Gaussians' colors, opacities and covariances to more accurately match the images. During training, adaptive density control will split, prune, and clone Gaussians ensuring there are just enough to faithfully represent the scene.

\subsection{Trajectory Optimization}

\subsubsection{Standard Methods}

Trajectory planning generates a path for a robot that is collision-free with the surrounding environment and further minimizes a performance metric such as execution time or control efforts.
Depending on the formulation, trajectory optimization also includes optimizing control inputs to follow the trajectory. Similar to other works~\cite{Chen2024,lei2024gaussnav,jin2024gs}, we assume robust low-level control that can track reasonably smooth state trajectories, so we omit the control inputs in our optimization formulation.

Graph-based methods such as A*~\cite{astar}, and sampling-based methods such as RRT*~\cite{karaman2011anytime} are straight-forward to implement and produce a series of collision-free waypoints. Trajectories produced by graph-based methods are often good initial guesses for other approaches. 
Unlike graph- or sampling-based methods, spline-based methods optimize a continuous trajectory instead of a series of waypoints. Collision avoidance is ensured by either constraining the convex hull of this spline to a predetermined safety corridor or by ensuring no collision at discretized points along the spline.
Zhou~et~al.~\cite{Zhou2020} obtain an obstacle gradient to push spline trajectories out of obstacles to achieve collision-free trajectories while Gao et al.~\cite{gao2016,Gao2019} construct a safe flight corridor around an RRT* path through a point cloud and optimize a spline trajectory inside the corridor to ensure safety.

\subsubsection{Radiance Field Methods}
There are numerous trajectory optimizing algorithms designed to work on standard scene representation such as meshes or point clouds and now  -- with the recent emergence of Radiance Fields-- \acp{nerf} are being used to tackle the problem.
Planning directly on Radiance Fields can be favorable as they are often trained from monocular images, and the extensive reconstruction allows for their use in multiple downstream tasks from planning and SLAM to teleoperation.
Pantic~et~al.~\cite{Pantic2022} propose a method to obtain an \ac{esdf} directly from a \ac{nerf} by inserting an additional head into the neural network to output the encoded distance. Adamkiewicz et al.~\cite{Adamkiewicz2021} directly use the opacity output by the \ac{nerf} as a collision measure at a set of points of the robot body. Starting from an A*-initial guess, a joint cost function consisting of a control effort and opacity term at the control points is minimized. 
In CATNIPS,~\cite{Chen2023} a probabilistic approach is used and shows that a \ac{nerf} is equivalent to a Poisson Point Process, which allows for direct quantification of the collision probability. Using this property, they formulate a chance constraint trajectory optimization.

Due to its novelty, few papers have been published on trajectory optimization for Gaussian Splats. In GaussNav~\cite{lei2024gaussnav}, Lei et al. reduce the Gaussian Splat to a point cloud of its means, which is voxelized and then projected to a two-dimensional cost map where trajectories are planned. This produces quick results but does not fully leverage the explicit 3D nature of \ac{3dgs}.
SplatNav~\cite{Chen2024} additionally takes the covariances of the Gaussians into account and computes a polyhedral safety corridor around an A* initial guess. The free space is determined by calculating the distance between the robot ellipsoid and the confidence intervals of the environment Gaussians. Inside this safety corridor, a spline trajectory is optimized to produce a safe path. 
The GS-Planner~\cite{jin2024gs} constrains the position along the trajectory to be outside of the $3\sigma$ confidence interval of the joint Gaussian of the ellipsoid and robot radius. Unlike SplatNav~\cite{Chen2024}, which uses a safety corridor as an intermediate representation, this constraint is directly imposed on the spline.
Goel and Tabib~\cite{goel2024distance} propose a method to derive an ESDF and collision probability of a Gaussian Surface Model for an ellipsoidal robot with a fixed orientation, which can then be used to plan trajectories. \\

While these works present considerable steps towards \ac{3dgs} planning, they consider only ellipsoidal robots with a fixed orientation.
However, in real world settings~\cite{dong2024efficient, Funk2023}, optimizing orientation is crucial to navigating through narrow passageways.
\textbf{By leveraging a detailed whole-body overlap model, we propose a solution to this shortcoming}.
Table~\ref{tab:comparison} provides an overview of how our algorithm compares to related work.

\section{Method}

Our methodology can be split into three parts:
\textit{1)} trajectory representation to create an initial spline, \textit{2)} collision measure and \textit{3)} optimization loop.

\subsection{Trajectory Representation}
We employ cubic B-splines~\cite{Holmer2022} for the optimization to generate a fully differentiable and smooth trajectories in $d$ dimensions.
With B-splines, every point along the trajectory can be expressed as a weighted sum of the $H$ control points $\mathbf{q}_i$ of the trajectory. Since we are using unclampled uniform cubic B-Splines, each spline segment is parameterized by $4$ control points, resulting in $H -3$ individual segments.
The progress along the trajectory is usually parameterized by a progress variable $s$, which can (but is not required to) be equal to the system time $t$. 
Using knots $\{-3, -2, \dots, H -1, H\}$ the spline is parametrized on $s \in [0, H - 3)$. Since each individual spline segment is parameterized between $[0,1]$, we define $s' = s - i$ with $i :=\lfloor s \rfloor$ to index the corresponding segment. The spline can then be evaluated with

\begin{equation}
\label{eq:point_wise_spline}
    \mathbf{x}(s') = 
\underbrace{
\begin{bmatrix}
1 & {s'} & {s'}^2 & {s'}^3
\end{bmatrix}
\frac{1}{6}
\begin{bmatrix}
1 & 4 & 1 & 0 \\
-3 & 0 & 3 & 0 \\
3 & -6 & 3 & 0 \\
-1 & 3 & -3 & 1
\end{bmatrix}
}_\Phi
\begin{bmatrix}
\mathbf{q}_{i+1}^T \\
\mathbf{q}_{i+2}^T \\
\mathbf{q}_{i +3}^T \\
\mathbf{q}_{i +4}^T
\end{bmatrix}
\end{equation}

By discretizing the trajectory with $\Delta s$ into $K$ discretization points and constructing a matrix $\underline{\Phi}$ from Equation \ref{eq:point_wise_spline} evaluated at each progress step, the relationship between discretized point matrix $\mathbf{X} = [\mathbf{x}_1\ \mathbf{x}_2\ \dots\ \mathbf{x}_K]^T \in \mathbb{R}^{K\times \text{d}}$ and control point matrix $\mathbf{Q} = [\mathbf{q}_1\ \mathbf{q}_1\ \dots\ \mathbf{q}_H]^T \in \mathbb{R}^{H \times \text{d}}$ is then given as

\begin{equation}
    \label{eq:spline_vectorized}
    \mathbf{X} =\underline{\Phi} \mathbf{Q}
\end{equation}

We use $\dot{\mathbf{X}}, \ddot{\mathbf{X}}, \dddot{\mathbf{X}} \in \mathbb{R}^{K\times \text{d}}$  consisting of discretizations $\dot{\mathbf{x}}_k$, $\ddot{\mathbf{x}}_k$, and $\dddot{\mathbf{x}}_k$ to denote the $k$-th discretized time derivatives of the spline.
Continous evaluations of the spline at $s$ and its time derivatives are denoted as $\mathbf{x}(s)$, $\dot{\mathbf{x}}(s)$, $\ddot{\mathbf{x}}(s)$,  and $\dddot{\mathbf{x}}(s)$.
Note that the spline is parameterized as a function of the progress variable $s$. We compute the time derivative with $\frac{d^i \mathbf{x}}{dt^i} = \frac{d^i \mathbf{x}}{ds^i} \left(\frac{ds}{dt}\right)^i$ and assume $m := \frac{ds}{dt} $ to be constant. 

This parametrization works well for simple vector spaces such as $\mathbb{R}^3$ but fails to interpolate between orientations. %
Spline representations for Lie Groups have been discussed in robotics literature~\cite{Sommer_2020}, which allow for pose optimization in $\text{SE}(3)$, encoding orientations.

Although this formulation is fully compatible with the proposed framework, we decided to follow a simpler parameterization because our target platform is a legged robot.
Instead of the full pose, \textbf{only the position $\mathbf{p} \in \mathbb{R}^3$ and yaw angle $\psi \in \mathbb{R}$ are optimized}. In the following, we assume $d = 4$ and decompose the trajectory evaluations as $\textbf{x} = [\mathbf{p}^T, \psi]^T \in \mathbb{R}^4$ , and control points $\mathbf{q}_i = [^{C}\mathbf{p}_i^T,  ^{C}\psi_i]^T\in \mathbb{R}^4$ with position component $^{C}\mathbf{p}_i^T$ and yaw component $ ^{C}\psi_i$. We use the $C$ superscript to denote a control point.

\subsection{Collision Measure}
To quantify the collision between two objects, we want to measure the overlap between them. The collision measure should, therefore, fulfill the following three properties:
\textit{1)} It should be close to zero if two objects do not overlap.
\textit{2)} It should be greater than zero when two objects overlap and quantify the magnitude of that overlap.
\textit{3)} It should be differentiable to allow for gradient evaluations in optimization.

In our work, we model both the environment $p(\mathbf{r})$ and the robot $r(\mathbf{r})$ by 3D Gaussian Splat.The environment is formed by $N$ individual Gaussians, and the robot by $M$ individual Gaussians.  We define the density of the robot and the environment at a specific point $\mathbf{r} \in \mathbb{R}^3$  in space as

\begin{equation}
    p(\mathbf{r}) = \sum_{i=1}^N\mathcal{N}(\mathbf{r};\bm{\mu}_i,\mathbf{\Sigma}_i) \in \mathbb{R}
\end{equation}
\begin{equation}
    r(\mathbf{r}) = \sum_{j=1}^M\mathcal{N}(\mathbf{r};\bm{\bar{\mu}}_i,\mathbf{\bar{\Sigma}}_j) \in \mathbb{R}
\end{equation}

The volume in which both quantities overlap is defined as the collision volume. To compute the magnitude of the collision, we compute the overlap integral between the environment field $p(\mathbf{r})$ and the robot field $r(\mathbf{r})$:

\begin{align}
\label{eq:convolution}
\int_{\mathbb{R}^3} p(\mathbf{r}) r(\mathbf{r}) dV &=   \sum_{i =1}^N \sum_{j=1}^M \int_{\mathbb{R}^3}\mathcal{N}(\mathbf{r};\bm{\mu}_i,\mathbf{\Sigma}_i)\mathcal{N}(\mathbf{r};\bm{\bar{\mu}}_j,\mathbf{\bar{\Sigma}}_j) dV\\
& =
\sum_{i =1}^N \sum_{j=1}^M \mathcal{N}(\bm{\bar{\mu}}_j;\bm{\mu}_i,\mathbf{\Sigma}_i + \mathbf{\bar{\Sigma}}_j)
\end{align}

Resulting in a representation that is fully differentiable with respect to the means $\bar{\bm{\mu}}_j$ and covariances $\mathbf{\bar{\Sigma}}_j$.
Note that these robot means and covariances can further be parameterized by, for example, the base pose and joint angles, whose gradients can then be derived with the chain rule.

This measure for collision fulfills the desired properties defined above.
Because of Gaussian's exponential nature, it converges to zero as the distance between the robot and the obstacle increase. That also means it does not contribute meaningfully to the  gradient at further distances.

Note that the quantity $\mathcal{N}(\bm{\bar{\mu}}_j;\bm{\mu}_i,\mathbf{\Sigma}_i + \mathbf{\bar{\Sigma}}_j)$ can be interpreted as the exponential of the negative Mahalanobis distance between $\bm{\bar{\mu}}_j$ and $\bm{\mu}_i$ using the joint covariance of the robot and environment Gaussian.

\begin{align}
\label{eq:mahalobis}    
\mathcal{N}
&= \text{exp}(- \frac{1}{2}(\bm{\bar{\mu}}_j - \bm{\mu}_i)^T(\mathbf{\Sigma}_i + \mathbf{\bar{\Sigma}}_j)^{-1} (\bm{\bar{\mu}}_j - \bm{\mu}_i)) \\
 & = \text{exp}(- \frac{1}{2}d_M(\bm{\bar{\mu}}_j, \bm{\mu}_i))
\end{align}
The Mahalanobis distance takes the potentially asymmetric extent of the Gaussians into account, and the exponential ensures the desired decaying property of the measure.

\subsection{Optimization Formulation}
Whenever integrating a quantity like the collision measure along the spline is necessary, we approximate it as a discrete sum over the values discretized along the spline in $K$ equidistant steps.
We represented the robot position and yaw orientation as a cubic B-spline.
The kinematics of each Gaussian that makes up the robot is a function of the position and orientation of the base $\bar{\bm{\mu}}_j(\mathbf{p}, \psi) = \bar{\bm{\mu}}_j(\mathbf{x})$.

We minimize the weighted sum of the obstacle cost, the jerk along the trajectory, and the distance of the final point to the goal with weights $\omega_1 = 0.1, \omega_2 = 40$ and $\omega_3 =1$.
The collision avoidance is included in the objective function instead of the constraints because the collision measure is not normalized, making finding an appropriate threshold unfeasible. 
A similar approach has been taken by~\cite{Adamkiewicz2021}, which faced a similar issue for \acp{nerf}.
We constrain the initial pose of the trajectory to the current position $\mathbf{x}_{\text{start}}$.
When planning for ANYmal, we further constrain the trajectory height to be at an appropriate distance $h$ above the ground so that the robot can follow it.

We ensure that the velocity and acceleration remain within the physical limits of mobile system. Instead of directly constraining samples of the respective derivatives of the spline along the trajectory, we leverage the convex hull property of splines.
Since each spline segment lies within the convex hull of its control points, it is enough to constrain the norm of the velocity and acceleration control points, to guarantee constraint satisfaction along the entire spline. However,  the conservativeness of this over approximation depends on the choice of spline basis. Because of this, we compute the convex hull with respect the \textbf{MINVO Basis}~\cite{tordesillas2022minvo}, which results in convex hulls of minimal volume, and therefore the least conservative approximation.
Linear and angular velocity and acceleration bounds are denoted as $v_{\text{max}}, a_{\text{max}}, \omega_{\text{max}}$, and $\alpha_{\text{max}}$ respectively. $\mathbf{x}_k^z$ denotes the $z$-component (height) of the $k$-th spline evaluation.

The optimization problem is defined in Equation \ref{eq:cost}.

\begin{align}
\label{eq:cost}
   \min_{\mathbf{Q}} \text{ } & \omega_1 \sum_{k=1}^K\sum_{i =1}^N \sum_{j=1}^M \mathcal{N}(\bm{\bar{\mu}}_j(\mathbf{x}_k);\bm{\mu}_i,\mathbf{\Sigma}_i + \mathbf{\bar{\Sigma}}_j) \\
     & + \omega_2 \sum_{k=1}^K\left \lVert \mathbf{\dddot{x}}_k \right \rVert + \omega_3 \lVert \mathbf{x}_K - \mathbf{x}_{\text{goal}} \rVert \\
    \text{s.t. } & \lVert \mathbf{x}_1 - \mathbf{x}_{\text{start}} \rVert = 0 \\
     & \lVert \mathbf{x}_k^z - h \rVert = 0 \text{ } \forall k \in \{1, \dots, K\}\\
     & \left \lVert \mathbf{\dot{p}}(s) \right \rVert \leq v_{\text{max}} \forall s \in [0, H - 3)\\
     & \left\lVert \mathbf{\ddot{p}}(s) \right \rVert \leq a_{\text{max}} \forall s \in [0, H - 3) \\
     & \left | \dot{\psi}(s) \right | \leq \omega_{\text{max}} \forall s \in [0, H - 3) \\
     & \left | \ddot{\psi}(s) \right | \leq \alpha_{\text{max}} \forall s \in [0, H - 3)
\end{align}

Due to the non-linear and constrained nature of the optimization problem, we employ an interior point method.
In order to create an initial guess, the \ac{3dgs} is used to create a rough occupancy grid, where a voxel is occupied if it contains at least one mean. A* is then run on this occupancy grid and used to initialize the control points by fitting a spline to the resulting path.

\subsection{Implementation Details}
We define and solve the described optimization problem with Casadi~\cite{andersson2019casadi}. We exploit the independent summation structure of the collision measure (Equation \ref{eq:convolution}) and run this computation on the GPU in parallel. This led us to define a custom Casadi functor that computes the evaluation of the function and is gradients in parallel using NVIDIA Warp~\cite{warp2022}.
Note that other GPU extensions for Casadi, such as L4Casadi~\cite{salzmann2023l4casadi} and CusADI~\cite{jeon2024cusadigpuparallelizationframework} exist, but were not directly applicable due to the custom 3D Gaussian overlap integral formulation.
The optimization problem is then solved via the interior point method (IPOPT)~\cite{wachter2006implementation} with the custom overlap integral functor.

\section{Experiments}
In this section, we evaluate our algorithm by applying it to planning problems in different environments represented by \ac{3dgs}.
All experiments are run on an Intel i7-8750H with 16 GB of RAM and an NVIDIA RTX 2070 Max-Q.

\subsection{Trajectory Evaluation}
To evaluate the functionality of our system, we plan trajectories through a range of \ac{3dgs} environments, both artificially generated scenes and ones created from real-world captures.
Furthermore, we demonstrate the use of the algorithm for navigating an ANYmal robot on hardware to highlight the importance of orientation-aware planning.

\subsubsection{ANYmal Navigation}
In order to plan through a \ac{3dgs} scene, the robot must be properly localized in a correctly scaled Gaussian environment. In deployments, this was accomplished by performing ICP to align accumulated LiDAR data with the surface means of the Gaussian Splat. In simulation, synthetic scenes can be created and arbitrarily scaled to provide complex testing scenes.

\begin{figure}[htbp]
    \centering
    \begin{subfigure}[b]{\linewidth}
        \centering
        \includegraphics[width=\linewidth]{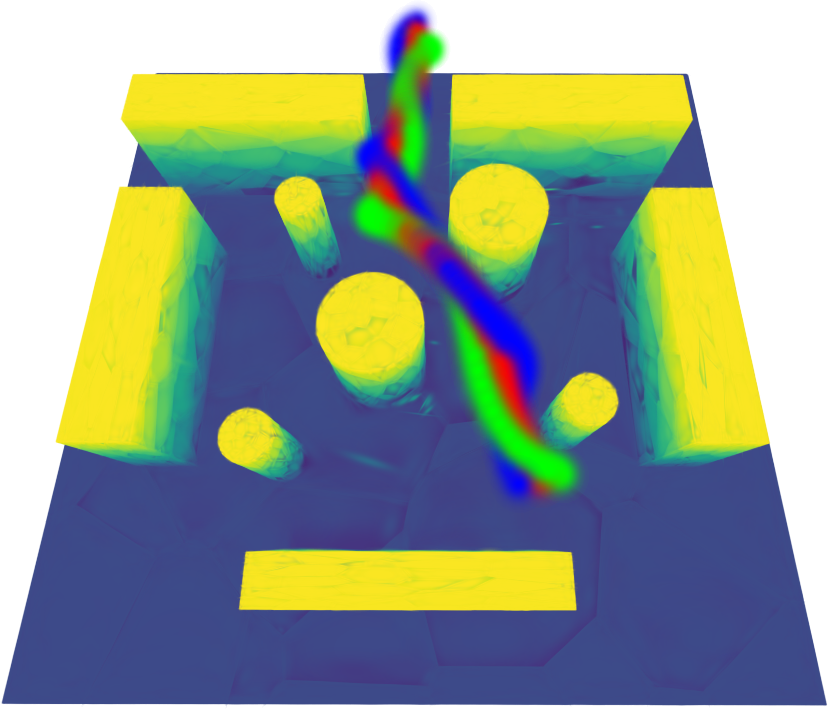}
        \caption{Pillar Environment}
        \label{fig:complex_small}
    \end{subfigure}
    \hfill
    \begin{subfigure}[b]{.9\linewidth}
        \centering
        \includegraphics[width=\linewidth]{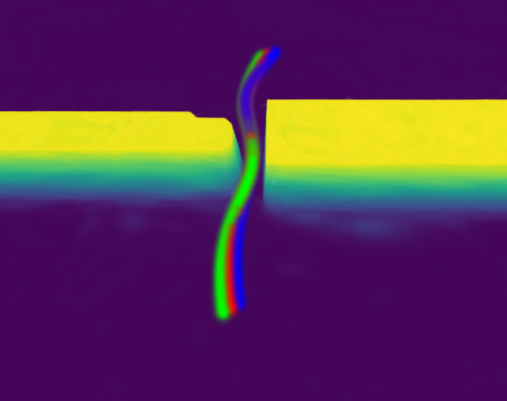}
        \caption{Narrow Corridor}
        \label{fig:narrow-corridor}
    \end{subfigure}
    
    \caption{Sample trajectories created on synthetic testing data showing a 3 Gaussian robot rotating to navigate the environments. The 3 Gaussians are shown in green for the front, red for the middle and blue for the rear of the robot, showing the rotation to navigate obstacles.}
    \label{fig:ANYmal_splats}
\end{figure}

\begin{table}[]
    \centering
    \begin{tabular}{c|c|c|c}
         Environment & Initial Guess Time & Solve Time & \# of Gaussians \\
         \hline
          Narrow Corridor  &  $0.24\si{\s}$   & $0.47\si{\s}$ &  24k \\
          Pillars          &  $0.25\si{\s}$  & $0.45\si{\s}$  &  49k \\
          \hline
          Machine Hall    &  $0.22\si{\s}$   &   $2.12\si{\s}$   &  243k \\
          Stonehenge      &  $0.55\si{\s}$   &   $0.83\si{\s}$   &  138k \\
    \end{tabular}
    \caption{Planning time for synthetic \ac{3dgs} scenes (top) and realistic scenes (bottom) using a 3 Gaussian robot.}
    \label{tab:runtime_envs}
\end{table}

Figure \ref{fig:complex_small} shows longer trajectories through the more complex pillar environment. The algorithm gracefully handles trajectories requiring additional turns while adapting the orientation to keep a maximum distance from the wall.

As Figure \ref{fig:narrow-corridor} shows, the planning algorithm effectively leverages the asymmetry of ANYmal to pass through the narrow opening collision-free. The start and end positions are constrained to be parallel to the wall; requiring the orientation to change to pass through the opening.

\subsubsection{General Trajectory Planning Through \ac{3dgs}}

\begin{figure}[htbp]
    \centering
    \begin{subfigure}[b]{\linewidth}
        \centering
        \includegraphics[width=\linewidth]{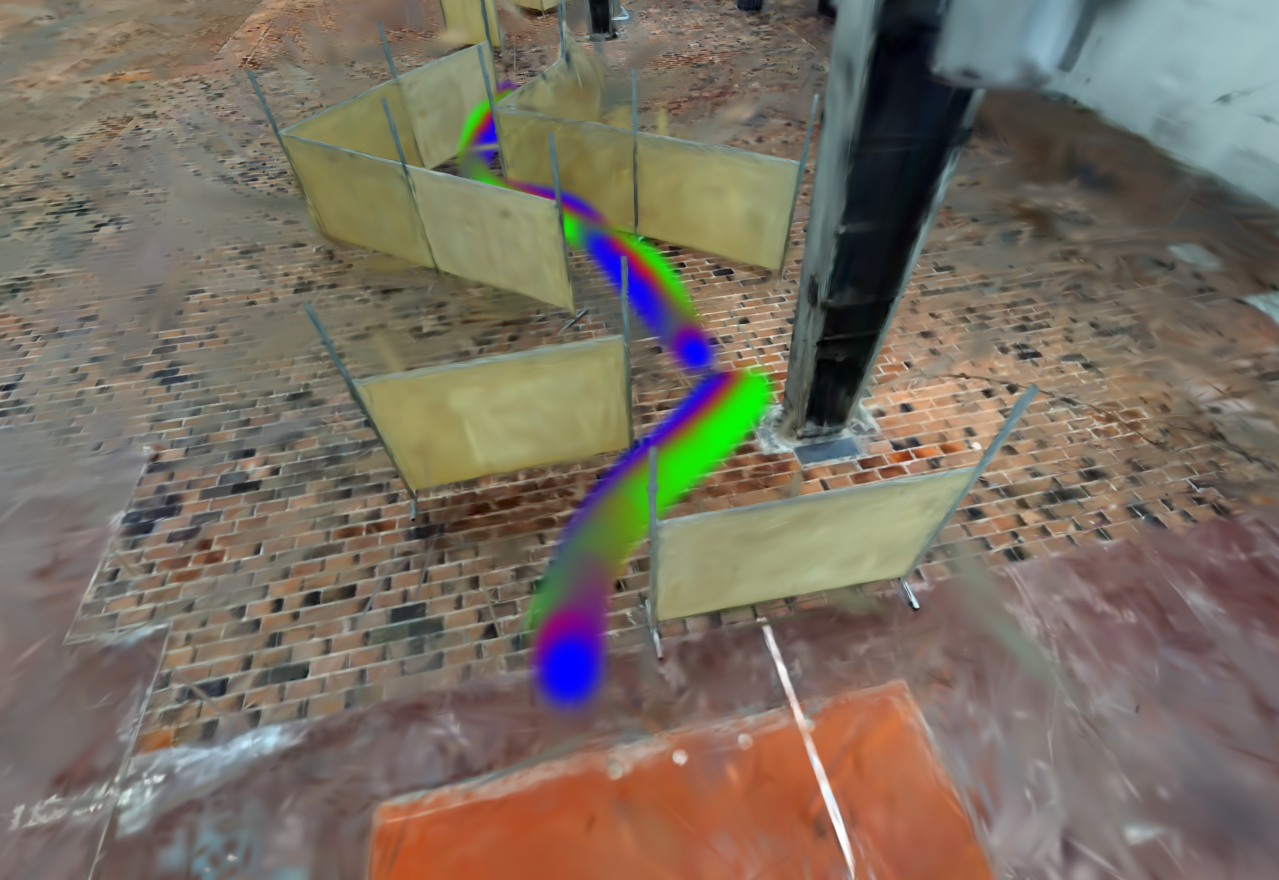} %
        \caption{Machine Hall}
        \label{fig:conference_room}
    \end{subfigure}
    \hfill
    \begin{subfigure}[b]{\linewidth}
        \centering
        \includegraphics[width=\linewidth]{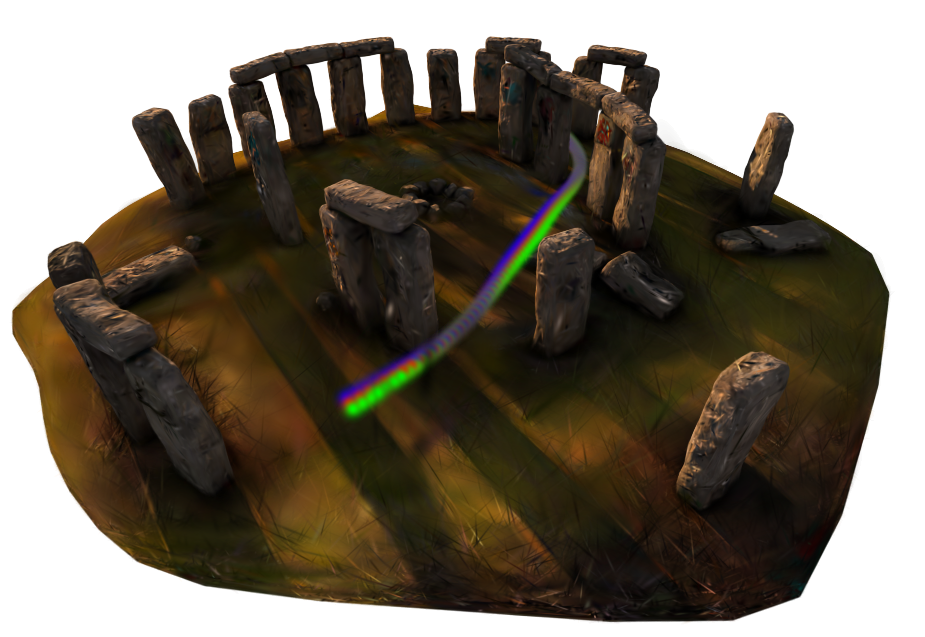} %
        \caption{Stonehenge Environment}
        \label{fig:rsl}
    \end{subfigure}

    \caption{Trajectories planned for a 3 Gaussian robot rotating through realistic scenes.}
    \label{fig:real_splats}
\end{figure}

Figure~\ref{fig:real_splats} shows that we can plan collision-free trajectories through splats that were created directly from the real-world environments. In the case of the machine hall sample, the data used for the reconstruction came entirely from the on-board sensors of the robot~\cite{wildersmith2024rfteleoperation}, quickly providing a scene with accurate scaling. 

\subsubsection{Hardware Experiments}

\begin{figure}[htbp]
    \centering
    \includegraphics[width=\linewidth]{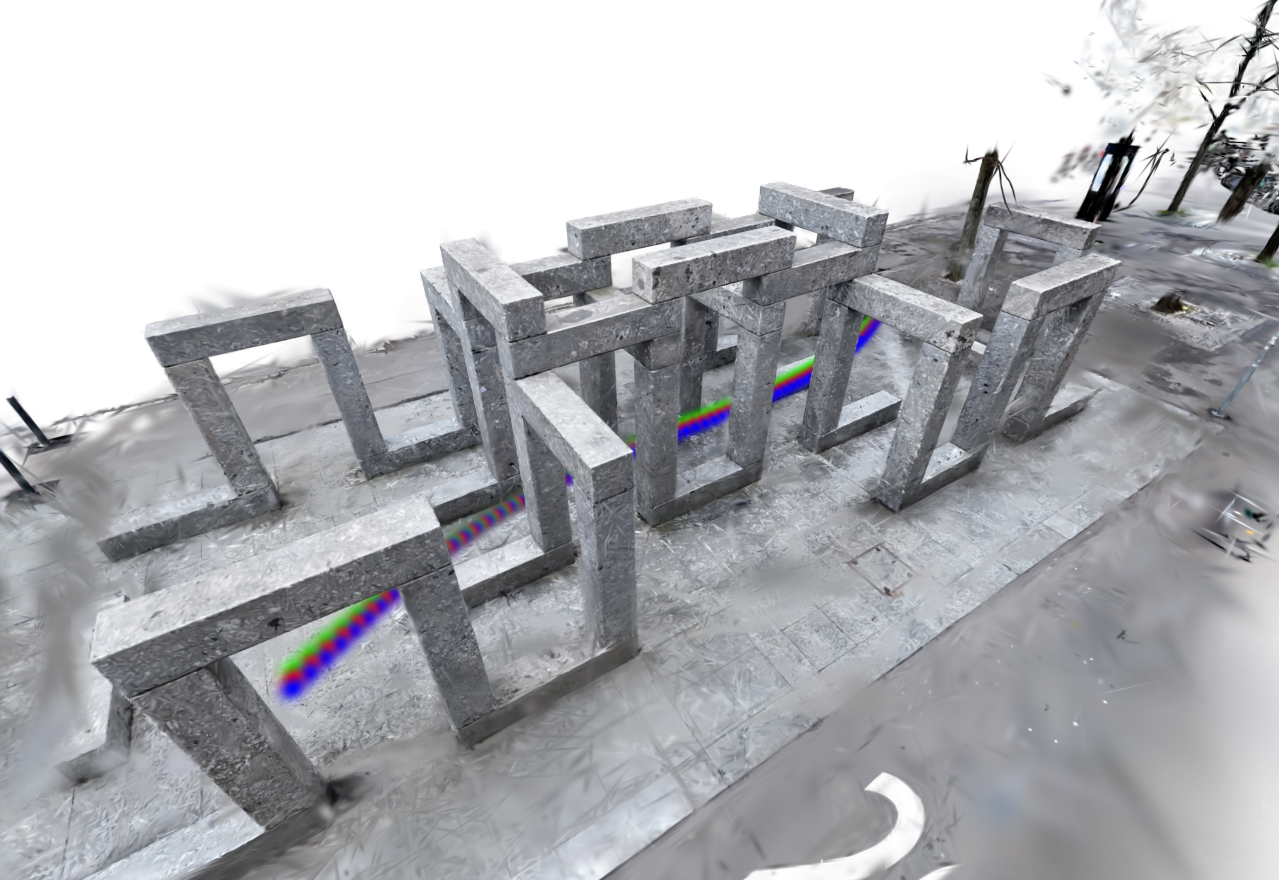}    
    \vspace{0.01cm}
    
    \includegraphics[width=\linewidth]{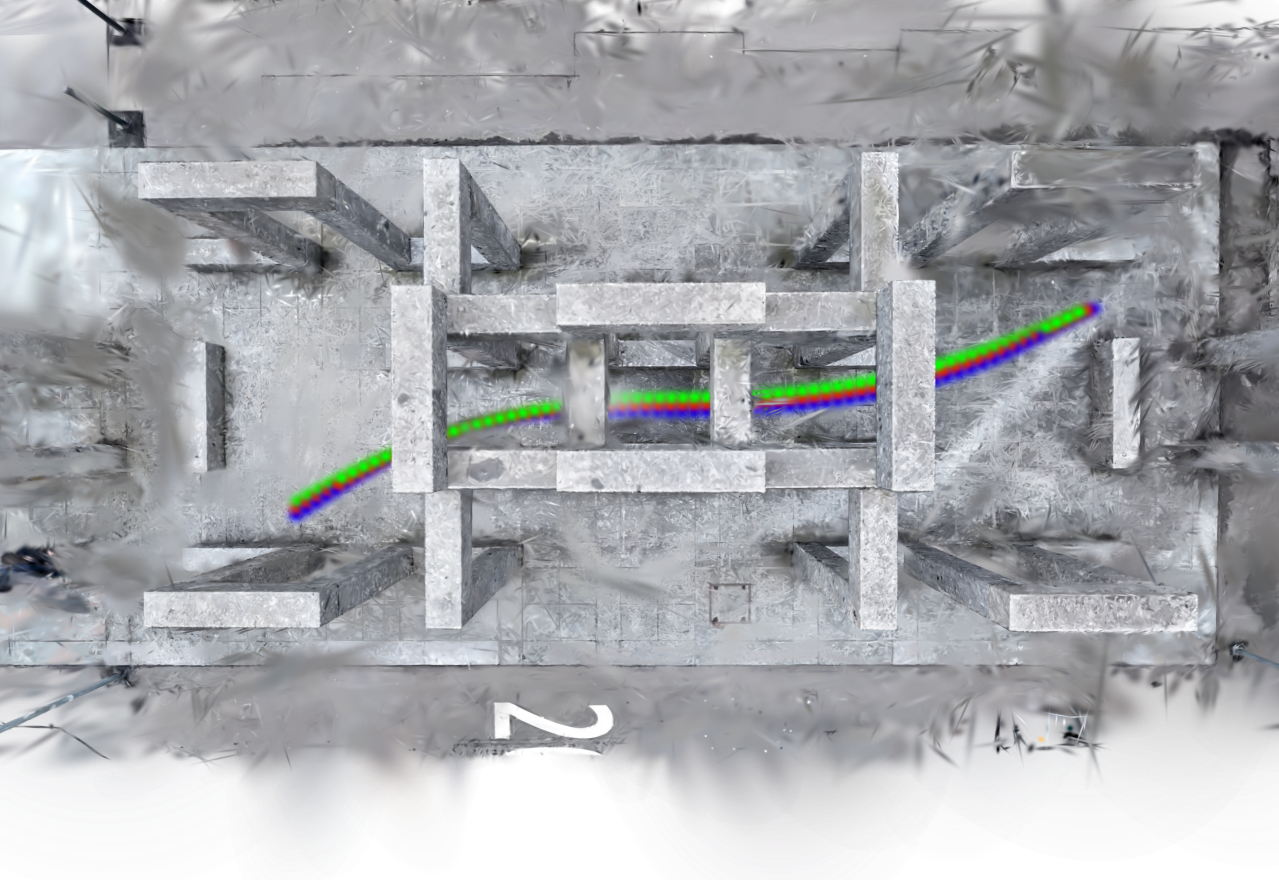}
       
    \includegraphics[width=\linewidth]{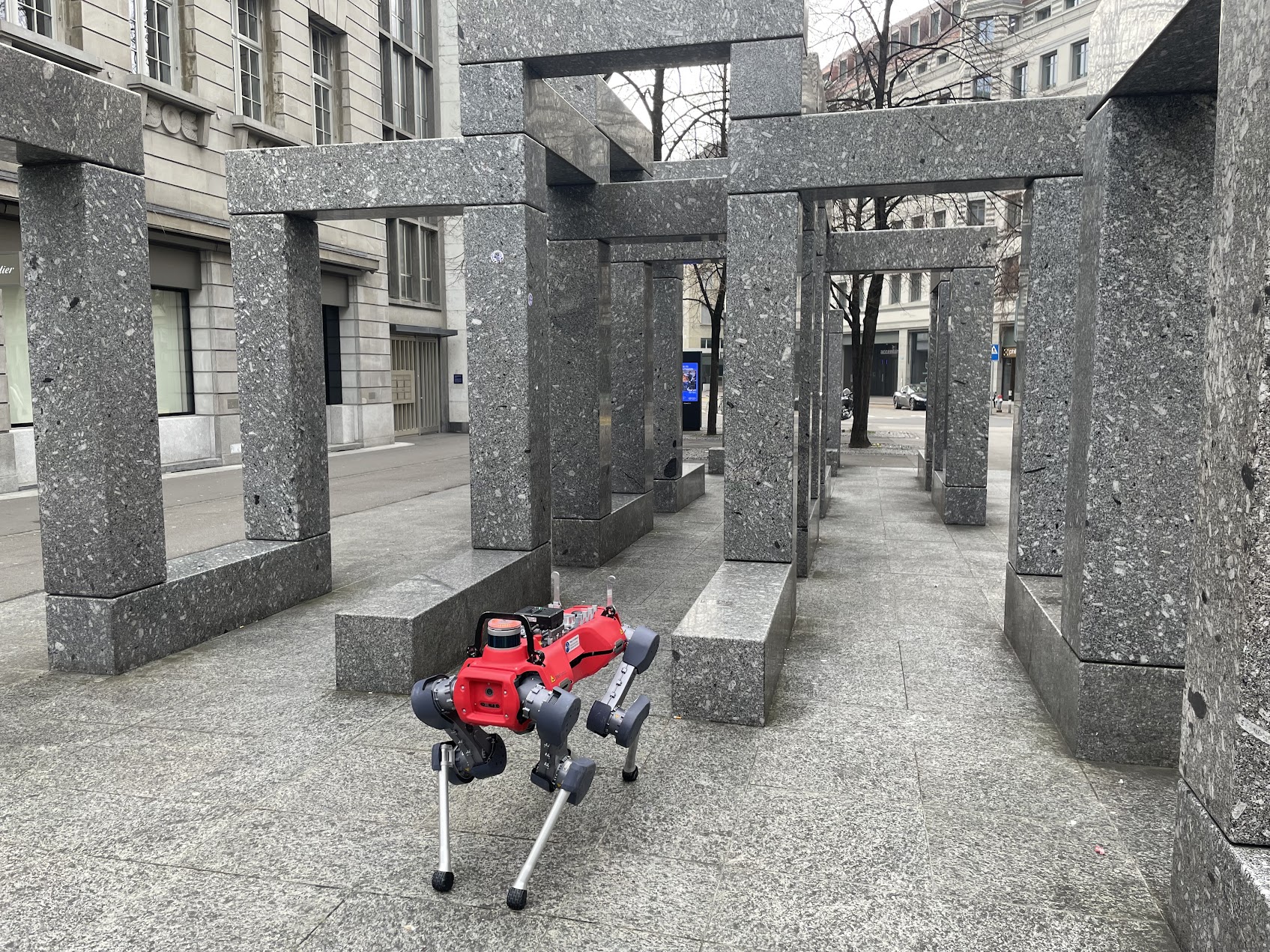}

\caption{Trajectories planned for the ANYmal robot through real world obstacles. On the bottom the actual robot can be seen at the site following the trajectories generated by our algorithm \textbf{FOCI}.}
    \label{fig:realworld}
\end{figure}

Finally, we showcase the method deployed on hardware using a real \ac{3dgs} reconstruction of a scene captured using drone imagery and localized on site using ICP on the LiDAR data and Gaussian means. The narrow passages visible in Figure \ref{fig:realworld} required the robot to rotate in order to safely traverse the environment, while obstacles on the ground forced the height-constrained trajectories to avoid shortcuts.

\subsection{Runtime}
We evaluate the performance of our method by comparing the runtimes of the Casadi optimization on a single CPU core, multiple CPU cores, and the GPU.
The computation time is generally linear with the number of environmental Gaussians, robot Gaussians, and collision discretization points.

\begin{figure}[htbp]
    \centering
    \includegraphics[width=\linewidth]{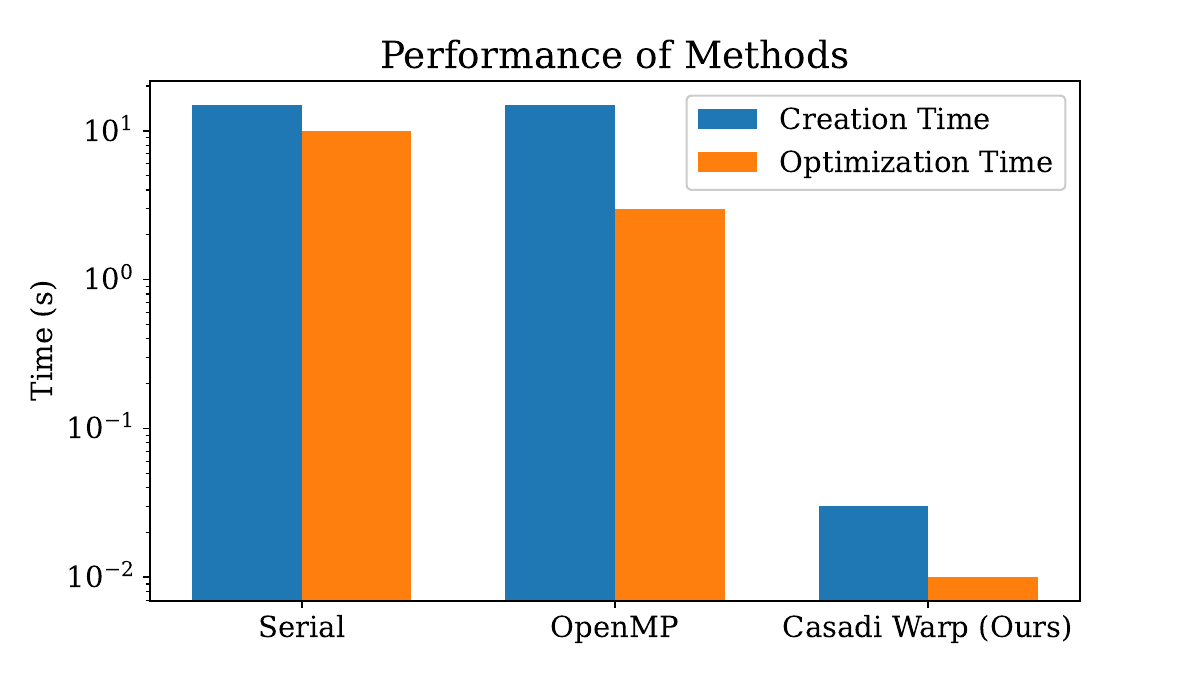}
    \caption{Comparison of the solver's creation and runtime running on the CPU and GPU for $50$k environmental Gaussians and one robot Gaussian. The ``serial" method is on a single CPU core, ``OpenMP" runs on multiple CPU cores and CasADi Warp is our custom GPU implementation.}
    \label{fig:runtimes}
\end{figure}

Figure \ref{fig:runtimes} shows the comparison between different computation methods.
While a significant speedup can be observed when optimizing on multiple CPU cores using OpenMP, \textbf{our custom GPU-enabled implementation is $320$ times faster}, often resulting in solutions that only take a few seconds.
\begin{figure}[htbp]

    \centering
    \includegraphics[width=\linewidth]{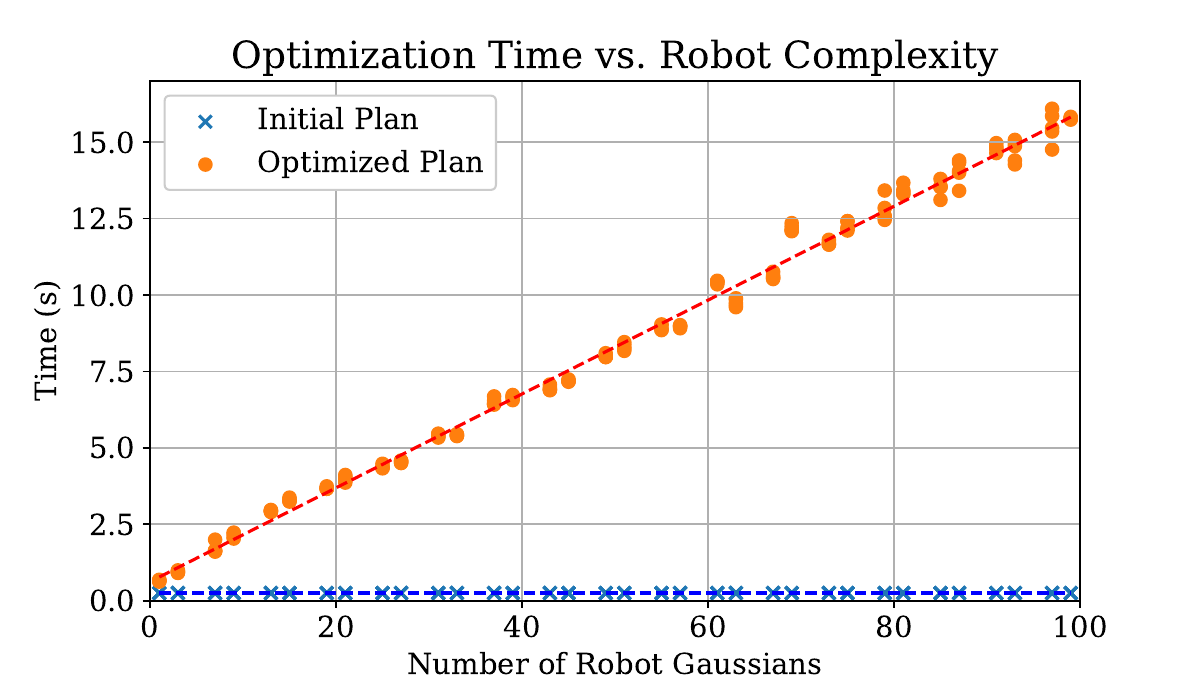}
    \caption{Optimization time for orientation-aware planning using increasingly complex robot models. These are planned through the Stonehenge environment with 138k Gaussians.}
    \label{fig:robot_complexity}
\end{figure}

\begin{figure}[htbp]

    \centering
    \includegraphics[width=\linewidth]{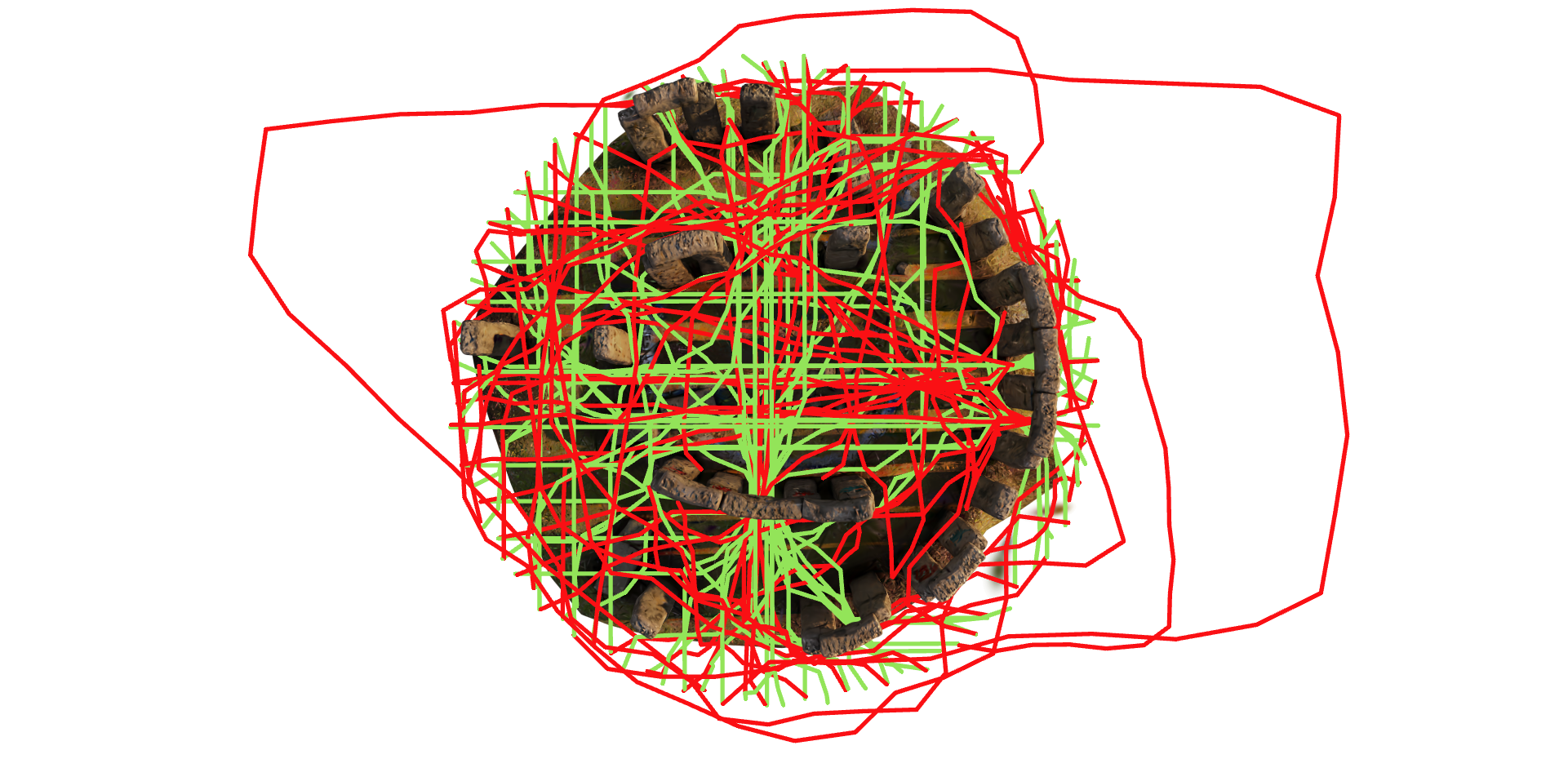}
    \caption{The collection of paths used to evaluate different methods with various start and end goals around Stonehenge. The RRT* path is is shown in \textcolor{red}{red} and Splat-Nav\cite{Chen2024} in \textcolor{green}{green}.}
    \label{fig:stonehenge_paths}
\end{figure}

\begin{table}[]
    \centering
    \begin{tabular}{c|c|c|c}
         Method & Solve Time (s) & Path Length (-) & Min. Safety Dist. (m) \\ 
         \hline
          RRT*          &  100.06   &   0.92    & 0.51 \\
          Splat-Nav     &  \textbf{0.76}   &   1.01    & 0.56 \\
          Ours          &  0.78   &   \textbf{0.88}    & \textbf{0.27} \\
    \end{tabular}
    \caption{Comparisons of the performance of our method with a similar \ac{3dgs} based planner and simple RRT* planner. Three metrics are used for evaluation, the speed of optimization, a path length relative to the scene scale, and minimum distance between the robot model and environment to measure the needed safety corridor.}
    \label{tab:baseline}
\end{table}

Table \ref{tab:runtime_envs} shows the total planning time for different environments of varying complexity. In the realistic environments, the larger scenes resulted in a longer A* search time, while the more complex 3 Gaussian robot slightly increases the general solve time. This relation between complexity of the robot model and solve time forms a linear relation as shown in Figure \ref{fig:robot_complexity}. 
In comparisons with similar methods (Table \ref{tab:baseline}, Figure \ref{fig:stonehenge_paths}) we are able to \textbf{surpass the speed of traditional methods such as RRT* on large complex scenes}, while having \textbf{similar time performance to state-of-the-art \ac{3dgs} methods such as Splat-Nav}. Additionally, our orientation-aware planning allows us to use halve the safety corridor distance in as we model a 1m by 0.5m robot such as ANYmal. Due to this unique ability to model the robot as a collection of Gaussians and therefore consider the robot's orientation, \textbf{the smaller safety corridor allowed for slightly shorter paths, leveraging the robot's geometry}. 

In Figure \ref{fig:robot_complexity} we compare the effect of more complex robot Gaussians on the solve time for large complex scenes such as Stonehenge. A \textbf{linear relation between robot complexity and optimization time} shows the potential of modeling more complex robot geometry, allowing even tighter safety corridors on more complex robots.

\section{Limitations}
Three current shortcomings of the algorithm include \textit{a)} occasional obstacle collision, \textit{b)} distance agnostic optimization, \textit{c)} sensitivity to \ac{3dgs} quality.

\begin{figure}[htbp]
    \centering
    
    \includegraphics[width=0.5\textwidth]{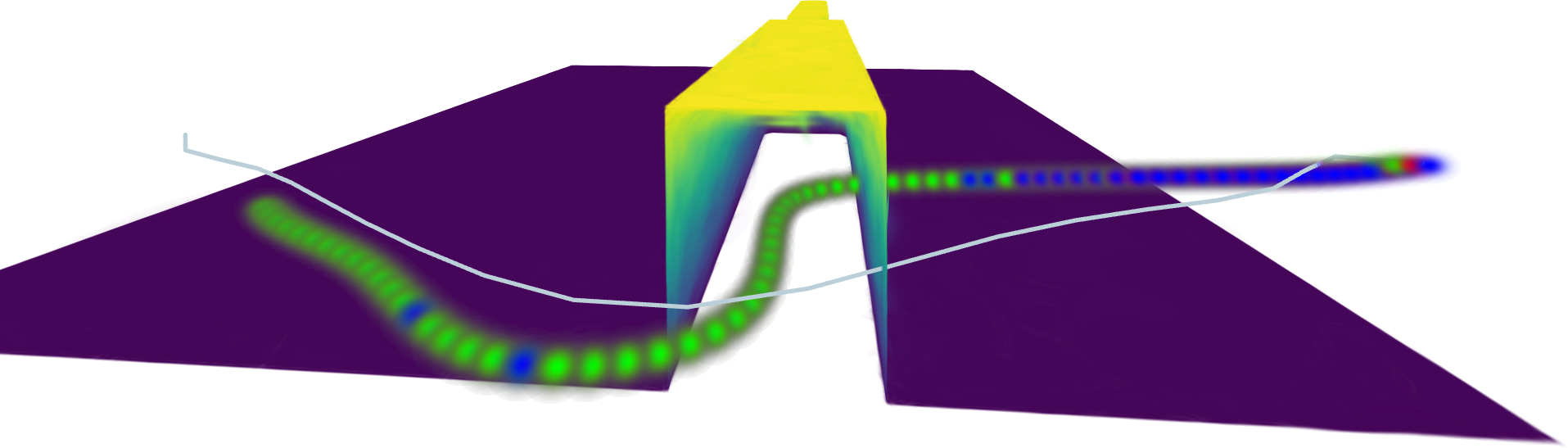}
    \caption{Optimized trajectory for which the collision avoidance fails.}
    \label{fig:collision_with_wall}
\end{figure}

\textit{a)} Both the jerk as well as the obstacle cost are additive terms in the cost function.
Since obstacle avoidance is not formulated as a hard constraint, it can be traded off with the jerk cost, yielding a low jerk but colliding trajectories. Figure \ref{fig:collision_with_wall} shows a trajectory resulting from such a trade off.
Related work, such as trajectory planning on NeRFs by Adamkiewicz et al.~\cite{Adamkiewicz2021}, has also encoded obstacle avoidance as a cost function component. We are confident that the existing issues can be resolved by constraining the trajectory to be close to the initial guess, tuning the weights of the individual cost terms, or using higher-order derivatives of the trajectory as the effort cost.

\textit{b)} Trajectories are parameterized over an interval that depends only on the number of control points.
Velocity, acceleration, and jerk are scaled with a constant time regulation factor $m$ that depends on the length of the initial guess.
Although this formulation is easy to implement, it has many impactful disadvantages.
Between two trajectories of the same shape but different lengths, the shorter one has a lower acceleration than the longer one because the robot has to be accelerated over larger distances while still being parameterized over the same progress interval. This means that longer trajectories automatically have bigger accelerations and jerks, since the optimization has no direct control over time.
Subsequently, the jerk cost can become the dominant term in the optimization, decreasing the importance of obstacle avoidance.
To account for different trajectory times, the execution time between the start and goal could be introduced into the trajectory representation and directly optimized as a decision variable.

\textit{c)} By using the overlap integral to compute collision between the robot and the scene, we assume that areas of high Gaussian overlap in the environment are dense objects geometrically, however this is not always the case. Instead, the \ac{3dgs} optimization process results in a high number of overlapping Gaussians in areas of high information density, both in terms of texture and geometric data. More complex shapes such as edges, fine strands, or lettering result in a large amount of Gaussians to accurately capture the geometry and texture. This means that when computing the overlap integral over the environment, flat regions with text or patterns have a slightly higher collision cost than clean flat surfaces. Additionally, as splatting only renders and optimizes Gaussians visible to the camera, the internal Gaussian density of objects is not guaranteed. This means that if the trajectory does collide with a wall it can get stuck in a local minima and not recover with the aid of internal collision gradients. Both of these can be addressed with more intelligent \ac{3dgs} creation, either by ensuring the Gaussians mainly represent geometry~\cite{chao2024texturedgaussiansenhanced3d}, or ensuring internal object density is consistent~\cite{goel2024distance}.

\section{Conclusion}
In this work, we proposed a novel collision formulation for \acl{3dgs}, that is computed in an efficient parallel manner, and integrated it into a trajectory optimization pipeline. 
We show that it can be effectively used for orientation-aware planning and verify it on the ANYmal quadruped robot.
Because we exclusively operate in 3D Gaussian space, representing the environment and the robot as Gaussians, our method can be freely combined with new developments from the 3DGS community. 
Furthermore, we show that this method works on realistic data including scenes captured using the onboard sensor of the robot itself.
The proposed method naively supports more complex robot kinematics, such as the kinematic chain of a robot arm. Future work can explore configuration-dependent trajectory planning for robotic manipulation tasks.
Since this method exclusively focuses on static and complete \acl{3dgs} environments, extensions toward dynamic scence and hybrid SLAM integration can be made in follow-up research.

\bibliographystyle{IEEEtranBST/IEEEtran}
\bibliography{references}

\onecolumn
\newpage

\end{document}